\documentclass[conference,letterpaper]{IEEEtran}
\IEEEoverridecommandlockouts
\usepackage{cite,float}
\usepackage{amsmath,amssymb,amsfonts}
\usepackage{algorithmic}
\usepackage{graphicx}
\usepackage{textcomp}
\usepackage{subfigure}
\usepackage{colortbl, booktabs}
\usepackage{xcolor}
\def\BibTeX{{\rm B\kern-.05em{\sc i\kern-.025em b}\kern-.08em
    T\kern-.1667em\lower.7ex\hbox{E}\kern-.125emX}}

\usepackage{url}

\definecolor{MyGreen}{HTML}{009900}
\definecolor{MyBlue}{HTML}{0d3983}
\RequirePackage[colorlinks,citecolor=MyGreen]{hyperref}
\usepackage{cleveref}
\usepackage{comment}
\usepackage{enumitem}
\usepackage{pifont}

\begin{document}

\title{Catch Causal Signals from Edges for Label Imbalance in Graph Classification
}

\author{

\IEEEauthorblockN{
Fengrui Zhang$^{1*}$\thanks{$^{*}$ Equal contribution.}
\qquad Yujia Yin$^{1*}$
\qquad Hongzong Li$^{2}$
\qquad Yifan Chen$^{1\dagger}$
\qquad Tianyi Qu$^{3, 4\dagger}$
}

\IEEEauthorblockA{
$^{1}$ Hong Kong Baptist University \qquad
$^{2}$ BayVax Biotech Limited \qquad
$^{3}$ Zhejiang University \qquad
$^{4}$ SF Tech \\
$^\dagger$~~Correspondence to: Yifan Chen \textlangle yifanc@hkbu.edu.hk\textrangle,
Tianyi Qu \textlangle qutianyi@sf-express.com\textrangle.
}
}

\maketitle

\begin{abstract}
Despite significant advancements in causal research on graphs and its application to cracking label imbalance, 
the role of edge features in detecting the causal effects within graphs has been largely overlooked, 
leaving existing methods with untapped potential for further performance gains. 
In this paper, we enhance the causal attention mechanism through effectively leveraging edge information to disentangle the causal subgraph from the original graph,
as well as further utilizing edge features to reshape graph representations. 
Capturing more comprehensive causal signals, 
our design leads to improved performance on graph classification tasks with label imbalance issues. 
We evaluate our approach on real-word datasets PTC, Tox21, and ogbg-molhiv, 
observing improvements over baselines. 
Overall, we highlight the importance of edge features in graph causal detection and provide a promising direction for addressing label imbalance challenges in graph-level tasks.
The preprint version is available at \url{https://arxiv.org/abs/2501.01707}. The model implementation details and the codes are available on \url{https://github.com/fengrui-z/ECAL}. 
\end{abstract}


\section{Introduction}

Graph Neural Networks (GNNs) have emerged as a powerful framework for learning from graph-structured data, deeply grounded in spectral graph theory and neural networks~\cite{scarselli2008graph,wu2020comprehensive}. They have been widely applied to graph classification and regression tasks, such as molecular property prediction~\cite{gnnmolecular,yang2019analyzing,duvenaud2015convolutional} and protein structure analysis~\cite{reau2023deeprank,jha2022prediction}. Their ability to capture intricate interactions between entities has proven highly effective. However, GNNs face considerable challenges in real-world scenarios involving distribution shifts, such as label imbalance \cite{wang2022imbalanced,zhao2021graphsmote,shi2020multi,juan2023ins}.

To address these challenges, researchers have extended GNNs to out-of-distribution (OOD) settings \cite{gui2022good,ju2024survey,li2022recent,zhang2024trustworthy}. One common approach involves environment-based methods \cite{yang2022learning,yuan2024environment}, which depend on prior knowledge of environments or require environment detection algorithms. Structured causal models offer an interpretable and flexible alternative for handling OOD data. For instance, researchers have combined environment adjustment with causal detection~\cite{gui2024joint,zhang2023out}, while others focus solely on causal modeling, introducing frameworks like causal inference~\cite{lin2024towards}, mutual information for identifying causal components~\cite{chen2022learning}, and re-weighting techniques to mitigate confounding effects~\cite{fan2023generalizing}. Recent advances, such as \textbf{causal attention mechanisms} and \textbf{structured causal models}, have further improved the understanding of causal relationships in graph-structured data. These mechanisms identify causal patterns by separating true causal features from shortcut features \cite{nauta2019causal, sui2022causal}, enabling better generalization in OOD scenarios.

Despite these advancements, many causal models overlook edge features, which are essential for capturing relational structures in graphs. Studies have demonstrated the potential of edge features in enhancing GNNs, including integrating edge information into GCNs~\cite{gong2019exploiting}, extending GATs to handle edge-featured graphs~\cite{hu2019strategies,wang2021egat,mo2022multi}, and incorporating edge encoding~\cite{ying2021transformers}.

In this paper, we enhance causal attention mechanisms by incorporating edge features. Specifically, we first reproduce the graph attention module with the capability to take edge features, and then further strengthen the design through learning from edge feature evolution (dubbed as EGATv1, EGATv2, respectively; c.f.\ \Cref{sec:efa}). These modules are integrated into the causal learning mechanism to detect causal subgraphs more effectively (corresponding to the ``attention estimation" step, c.f.\ \Cref{sec:prelim}) and accordingly strengthen the output graph representation. In summary,

\begin{itemize}[leftmargin=*]
    \item We recognize that causal models on graphs usually overlook the causal signals from edges.
    \item We introduce two edge-enhanced modules to tweak the causal attention mechanism, which leads to a better detection of causal structures in graphs .
    \item Extensive experiments are performed on label-imbalance graph classification tasks, which verifies the efficacy of incorporating edge features into causal models.
\end{itemize}

\begin{figure}[t]
    \centering
    \includegraphics[width=1\linewidth]{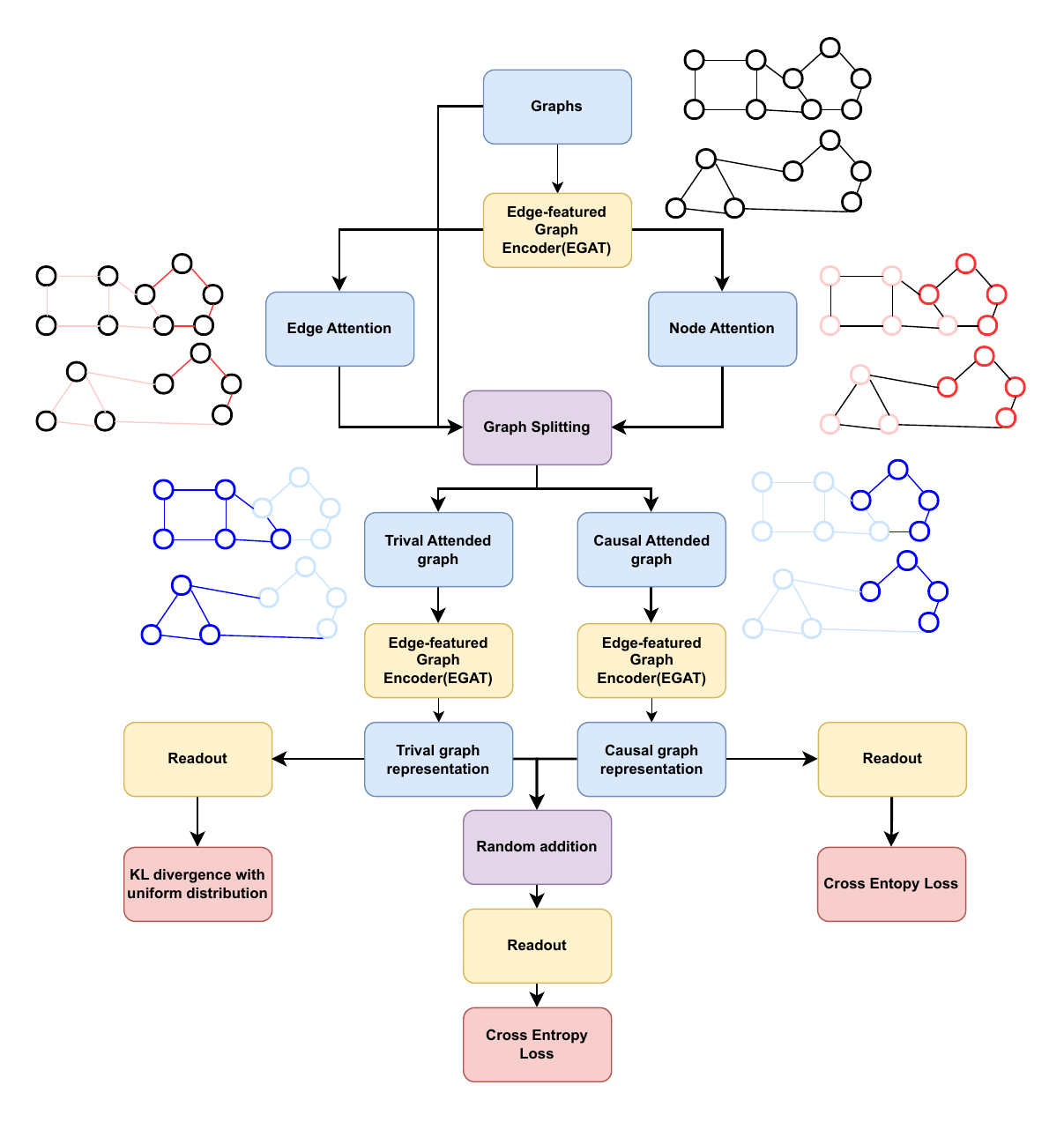}
    \caption{Framework overview. The proposed model integrates causal attention mechanisms with edge feature incorporation, enhancing the generalization capabilities of GNNs in OOD scenarios.
    Best viewed zoom in.}
    \label{fig:framework}
\end{figure}

\section{Preliminaries}
\label{sec:prelim}

\subsection{Task formulation}
For a general graph $G = (\mathcal V, \mathcal E)$ composed of the node set~$\mathcal V$ and the edge collection $\mathcal E$, 
we represent it as a triplet of matrices \((\mathbf A, \mathbf X, \mathbf E) \), where:
\begin{itemize}[leftmargin=*]
    \item \( \mathbf A \in \mathbb{R}^{|\mathcal V| \times |\mathcal V|} \) is the adjacency matrix indicating the connectivity between the nodes.
    \item \( \mathbf X \in \mathbb{R}^{|\mathcal V| \times d_v} \) is the node feature matrix, whose rows $X_i$'s correspond to \( d_v \)-dimensional feature vectors for each node.
    \item \( \mathbf E \in \mathbb{R}^{|\mathcal E| \times d_e} \) is the edge feature matrix, where for $(i, j) \in \mathcal E$, \(\mathbf{e}_{ij} \in \mathbb{R}^{d_e}\) denotes the \(d_e\)-dimensional feature vector for the edge connecting nodes \(i\) and \(j\) in the graph.
\end{itemize}
The task is to learn a function \( f: G \mapsto y \), where \(y \in \mathcal{Y} \) is one of the possible graph labels. 
Due to distributional shifts, the distribution of a graph $G$ in the testing set usually differs from the one in training.

\subsection{Causal discovery and causal attention mechanisms}

\emph{Causal discovery} aims to identify the causal components in complex data, such as graphs, by distinguishing causality from spurious correlations. Attention mechanisms have proven effective in this domain by selectively focusing on relevant nodes and edges, enabling more accurate inference.

A typical framework for causal attention mechanism~\cite{sui2022causal} is composed of two steps.
\ding{182}~\emph{Attention estimation}: the framework will first dynamically adjust the attention score to locate the most relevant nodes/edges and thus differentiate between causal and spurious correlations;
as an illustration, this step mainly corresponds to the top yellow ``EGAT'' box in \Cref{fig:framework}.
\ding{183}~\emph{Graph representation}: attaining causal and non-causal subgraphs, the framework will then further apply attention modules to extract features;
this step is indicated by the bottom two yellow ``EGAT'' box in \Cref{fig:framework}.

\section{Methods}

In this work, we aim to address the challenge of label-imbalance graph classification by incorporating edge features into a causal attention mechanism. 
Our proposed framework is illustrated in Figure~\ref{fig:framework}, highlighting the process from edge-featured graph encoding/splitting to the training loss.

The process starts by passing the input graph through an edge-featured graph encoder to extract node and edge causality. Edge and node attention scores are estimated (step \ding{182}, c.f.\ \Cref{sec:efa}) to identify significant components. As described in \Cref{sec:loss}, these scores split the graph into a causal subgraph with key elements and a trivial subgraph with less relevant components. Both subgraphs are processed by separate encoders (step \ding{183}). The training loss includes three parts: cross-entropy loss for the causal subgraph, KL divergence for uniformity of the trivial subgraph, and cross-entropy loss after randomly pairing causal and trivial subgraphs within the same batch.

\subsection{Incorporating edge features into attention}\label{sec:efa}

Regarding the integration of edge features into the attention mechanism, it is intuitive to use the Edge-Featured Graph Attention Network (EGAT) \cite{wang2021egat}.
In this module, the causality attention score \( \alpha_{i} \) for each node \( i \) and the causality attention score \( \alpha_{ij} \) for each edge \( (i, j) \in \mathcal E \) (i.e., node~$j$ must be within $\mathcal{N}(i)$, the neighborhood of node $i$) are calculated considering the information from the node and the edge. 
(We note this module, referred to as EGATv1, will shortly be enhanced and ultimately not employed in step~\ding{182}; however, this module will be evaluated in step \ding{183} graph representation.)

The node features \( X_i, X_j \) and the edge feature \( \mathbf{e}_{ij} \) are first transformed through linear projections; %
the concatenation of the transformed features can be expressed as:
\[
\left[ X_i' \parallel X_j' \parallel  \mathbf{e}_{ij}'\right],
\]
where \( X_i' = W_x^{(i)} X_i \), \( X_j' = W_x^{(j)} X_j \), and \( \mathbf{e}_{ij}' = W_e \mathbf{e}_{ij} \).
The node embedding from the EGAT is computed as 
\begin{equation}
X_i^{out} = \text{MLP} \left( \sum_{j \in \mathcal{N}(i)} a_{ij} X_j' \right)
\end{equation}
where the attention score $a_{ij}$ involves the comprehensive information from $\left[ X_i' \parallel X_j' \parallel  \mathbf{e}_{ij}'\right]$, computed as
\[
a_{ij} = \frac{\exp \left( \text{\scriptsize LeakyReLU} \left( a^T \left[X_i' \parallel X_j' \parallel \mathbf{e}_{ij}' \right] \right) \right)}{\sum_{k \in \mathcal{N}(i)} \exp \left(\text{\scriptsize LeakyReLU} \left( a^T \left[X_i' \parallel X_k' \parallel \mathbf{e}_{ik}' \right] \right) \right)},
\]
where $a$ indicates learnable parameters.
Then we adopt two MLPs to estimate the node causality  attention scores $\alpha_{i}$ and the edge causality attention scores $\alpha_{ij}$ respectively based on $X_i^{out}$ and $\mathbf{e}_{ij}'$: 
\begin{align}
    \alpha_{i},1-\alpha_{i}&=\sigma(\mathrm{MLP}_{\mathrm{node}}(X_i^{out}))\\
\alpha_{ij},1-\alpha_{ij}&=\sigma(\mathrm{MLP}_{\mathrm{edge}}(\mathbf{e}_{ij}'))
\label{eqn:edge-score}
\end{align}
where $\sigma(\cdot)$ is the softmax function.

\textbf{An enhanced EGAT alternative}:
we note the edge causality attention scores in \Cref{eqn:edge-score} are computed using edge information $\mathbf{e}_{ij}'$, which is isolated from the neighborhood information.
To better leverage edge features, we use another EGAT design~\cite{kaminski2022rossmann}, and we call it EGATv2 to distinguish it from the design above. 
This structure updates both edge and node features, allowing for a more effective capture of the dynamic relationships between nodes;
in EGATv2, given the feature $\left[ X_i' \parallel X_j' \parallel  \mathbf{e}_{ij}'\right]$, the edge embedding is first calculated as ($W_e^{out}$ indicates learnable parameters):
\[
\mathbf{e}_{ij}^{out} = \text{\scriptsize LeakyReLU} \left( W_e^{out} \left[X_i' \parallel \mathbf{e}_{ij}' \parallel X_j' \right] \right).
\]
The node embedding is next generated based on $\mathbf{e}_{ij}^{out}$:
\begin{equation}
X_i^{out} = \text{MLP} \left( \sum_{j \in \mathcal{N}(i)} a_{ij}\left(\mathbf{e}_{ij}^{out}\right) X_j' \right),
\end{equation}
where $a_{ij}$ is now a function of $\mathbf{e}_{ij}^{out}$ and reads
\[
a_{ij} = \frac{\exp \left(a^T \mathbf{e}_{ij}^{out}\right)}{\sum_{k \in \mathcal{N}(i) } \exp \left(a^T \mathbf{e}_{ik}^{out} \right)}.
\]
We then follow a process akin to EGATv1, utilizing two MLPs to estimate the node causality attention scores and edge causality attention scores, with $X_i^{out}, \mathbf{e}_{ij}^{out}$ as inputs this time: 
\begin{align*}
    \alpha_{i},1-\alpha_{i}&=\sigma(\mathrm{MLP}_{\mathrm{node}}(X_i^{out}))\\
\alpha_{ij},1-\alpha_{ij}&=\sigma(\mathrm{MLP}_{\mathrm{edge}}(\mathbf{e}_{ij}^{out}))
\end{align*}
where $\sigma(\cdot)$ is the softmax function. 
Compared to EGATv1, 
the edge causality attention scores are now estimated by the more comprehensive edge embedding $\mathbf{e}_{ij}^{out}$.
We term this design (with EGATv2) for step \ding{182} attention estimation as edge-enhanced causal attention learning (\textbf{ECAL}) along this paper.

\subsection{Graph causal representation with edge information}
\label{sec:loss}

The core idea within our approach is to enhance causal discovery by incorporating edge features to improve OOD generalization in graph-level classification. In the following, we propose the complete design to create better representations of causal structures by leveraging both node and edge features.

\subsubsection{Causal subgraph splitting} %
With the causality attention scores estimated in~\Cref{sec:efa}, we are now able to generate two subgraphs, the \textbf{causal subgraph} \( G_c \) and the \textbf{trivial subgraph} \( G_t \), from the original graph $G$:
\begin{align}
    G_c=(\mathbf A, M_x \cdot \mathbf X, M_e \cdot \mathbf E ),\\
    G_t=(\mathbf A, \bar{M_x} \cdot \mathbf X , \bar{M_e} \cdot \mathbf E),
\end{align}
where $M_x=\text{diag}(\alpha_{i})$, $M_e=\text{diag}(\alpha_{ij})$ will filter out the non-causal nodes and edges; 
accordingly, their ``complement'' $\bar{M_x}=I_x-M_x$, $\bar{M_e}=I_e-M_e$, where $I_x$ and $I_e$ are the identity matrix.
As a result, nodes and edges with higher causality attention scores receive larger weights in \( G_c \), emphasizing their relevance to the task. 
Conversely, those with lower attention values are weighted less in \( G_t \), reflecting their diminished importance for the causal task.

\subsubsection{Edge-featured causal and trivial representations}

For the causal subgraph \( G_c \) and the trivial subgraph \( G_t \), we compute their representations by incorporating both node and edge features, allowing stronger ability to capture significant causal relationships while distinguishing them from trivial ones. 

To simplify the notations, in this part the causal and the trivial subgraphs are referred to as:
\begin{equation}
G_c = \{\mathbf A, \mathbf X^{(c)}, \mathbf E^{(c)} \}, \quad G_t = \{\mathbf A, \mathbf X^{(t)}, \mathbf E^{(t)} \}.
\end{equation}
We reuse the EGAT modules (both v1 and v2 are evaluated in the experiments, c.f.\ the experiments in \Cref{tab:my-table}) to obtain the representations of the attended graphs. 
The final label prediction $\hat{y}^{(x)}$ is obtained via a readout function $f_{\text{readout}}^{(x)}(\cdot)$ and a classifier $\Phi_x$ (the script $x$ can be $c$ or $t$, corresponding to ``causal'' or ``trivial''), which can be formulated as :
\begin{align*}
\hat{y}^{(c)} &= \Phi_c(h^{(c)}), \text{with~} h^{(c)} = f_{\text{readout}}^{(c)}(\text{EGAT}_{c}(\mathbf X^{(c)}, \mathbf E^{(c)})),
\\
\hat{y}^{(t)} &= \Phi_t(h^{(t)}), \text{with~} h^{(t)} = f_{\text{readout}}^{(t)}(\text{EGAT}_{t}(\mathbf X^{(t)}, \mathbf E^{(t)})).    
\end{align*}
This approach ensures that the model focuses on the most relevant parts of the graph while down-weighting less informative features, ultimately leading to better generalization in out-of-distribution (OOD) graph classification tasks.

\begin{table*}[!t]
\centering
\caption{Test Accuracy in graph classification tasks. 
The best accuracy results are boldfaced.}
\label{tab:my-table}
\begin{tabular}{cccccccc}
\hline
Dataset     & PTC-FM         & PTC-MM         & Tox21-AR       & Tox21-ATAD5    & Tox21-PPAR-gamma & Tox21-aromatase & ogbg-molhiv   \\ \hline

GCN         & 0.457          & 0.559          & 0.981          & 0.992          & 0.984            & 0.932           & 0.801         \\ 

\rowcolor{gray!30} GAT         & 0.400          & 0.485          & 0.989          & 0.990          & 0.993            & 0.935           & 0.796         \\ 

EGATv1      & 0.386          & 0.485          & 0.994          & 0.988          & 0.987            & 0.937           & 0.806         \\ 

\rowcolor{gray!30}

EGATv2      & 0.443          & 0.515          & 0.99           & 0.988          & 0.984            & 0.932           & 0.793         \\

CAL+GCN     & 0.429          & 0.574          & 0.985          & 0.989          & 0.983            & 0.943           & 0.801         \\ 

\rowcolor{gray!30}

CAL+GAT     & 0.429          & 0.456          & 0.99           & 0.988          & 0.98             & 0.931           & 0.806         \\ 

ECAL+EGATv1 & 0.386          & \textbf{0.691} & 0.994          & 0.982          & 0.989            & 0.921           & \textbf{0.85} \\ 

\rowcolor{gray!30}

ECAL+EGATv2 & \textbf{0.671} & \textbf{0.691} & \textbf{0.996} & \textbf{0.995} & \textbf{0.997}   & \textbf{0.947}  & \textbf{0.81}\\
\hline
\end{tabular}
\end{table*}

\begin{table*}[!t]
\centering
\caption{$p$-values in two-sample $t$-tests for comparing ECAL with CAL.}
\label{tab:p-value-table}
\begin{tabular}{|c|c|c|c|c|c|c|c|}
\hline
Dataset     & PTC-FM & PTC-MM & Tox21-AR & Tox21-ATAD5 & Tox21-PPAR-gamma & Tox21-aromatase & ogbg-molhiv \\ \hline
p-value     & 3.9e-06 & 2.9e-08 & 7.4e-12   & 3.7e-4       & 3.3e-05           & 0.018            & 7.3e-14     \\ \hline
\end{tabular}
\end{table*}

\subsubsection{Loss computation}

We train the model using both the causal subgraph \( G_c \) and the trivial subgraph \( G_t \). 
The supervised label corresponding to the causal subgraph should be the ground-truth one, and we apply the cross-entropy function to obtain the classification loss:
\begin{equation}
\mathcal{L}_{\text{CE}} = - \sum_{k} y_k^T \log \hat{y}^{(c)}_k
\end{equation}
where \( y_k \) is the one-hot label vector for the $k^{th}$ sample.

In addition, to make causality detection feasible, two more loss terms are involved.
For the trivial subgraph, we apply a KL divergence loss to push these predictions \( \hat{y}^{(t)} \) towards a uniform distribution \( \mathcal{U} \), ensuring that less important elements contribute minimally:
\begin{equation}
\mathcal{L}_{\text{KL}} = \sum_{k}\text{KL} \left(  \hat{y}^{(t)}_k \parallel \mathcal{U} \right)
\label{eqn:loss-kl}
\end{equation}
Furthermore, guided by the spirit of intervention in backdoor adjustment~\cite{pearl2009causality}, the random addition technique is applied to isolate the effectiveness of the causal subgraph from the trivial subgraph on the representation level~\cite{correa2017causal,correa2018generalized}. 
In more details, in this step we randomly pair the causal graph with a mismatched trivial graph from the same training batch, and compute a separate cross-entropy loss using the predictions from the randomized representation \( \hat{y}_{(k,k')} = \Phi(h^{(c)}_k+h^{(t)}_{k'}) \) ($\Phi(\cdot)$ is a third classifier) and the target output:
\begin{equation}
\mathcal{L}_{\text{BA}} = - \sum_{k}\sum_{k'} y_k^T \log \hat{y}_{(k,k')}
\label{eqn:loss-ba}
\end{equation}
The total training loss is a weighted combination of the three aforementioned loss terms:
\begin{equation}
\mathcal{L} = \mathcal{L}_{\text{CE}} + \mathcal{L}_{\text{CD}}(\lambda_1,\lambda_2),
\end{equation}
where $\mathcal{L}_{\text{CD}}(\lambda_1,\lambda_2)=\lambda_{1} \mathcal{L}_{\text{KL}} + \lambda_{2} \mathcal{L}_{\text{BA}}$ represents the loss for causality detection.
This strategy ensures that the model focuses on learning from the causal subgraph while minimizing the influence of trivial features.

\section{Experiment Results}

We conducted experiments on three edge-featured graph classification datasets, including Tox21~\cite{mayr2016deeptox,huang2016tox21challenge}, ogbg-molhiv~\cite{hu2020open}, and PTC-MR~\cite{bai2019unsupervised}. 
We manipulate the dataset to introduce label imbalance so that the distribution of training datasets and that of testing datasets are different.

\subsection{Baselines and main comparison}
To illustrate the superiority of the proposed edge-enhanced framework for the causal attention mechanism, we employ three relevant frameworks for graph classification as baselines:
\begin{itemize}[leftmargin=*]
    \item \textbf{General models:} GCN\hspace{0pt}\cite{kipf2016semi}, GAT\hspace{0pt}\cite{velivckovic2017graph}
    \item \textbf{General models + edge info:} EGATv1\hspace{0pt}\cite{wang2021egat}, EGATv2\hspace{0pt}\cite{kaminski2022rossmann}
    \item \textbf{OOD models:} CAL+GCN\hspace{0pt}\cite{sui2022causal}, CAL+GAT\hspace{0pt}\cite{sui2022causal}
\end{itemize}
The evaluation metric is the classification accuracy, and the results are summarized in~\Cref{tab:my-table}. To assess statistical significance, $p$-values calculated from 10 independent trials are shown in~\Cref{tab:p-value-table}, confirming the consistent improvements achieved by ECAL over CAL.

\subsection{Ablation studies}
\subsubsection{Ablating causality detection}
\begin{figure}[t!]
    \centering
    \includegraphics[width=0.9\linewidth]{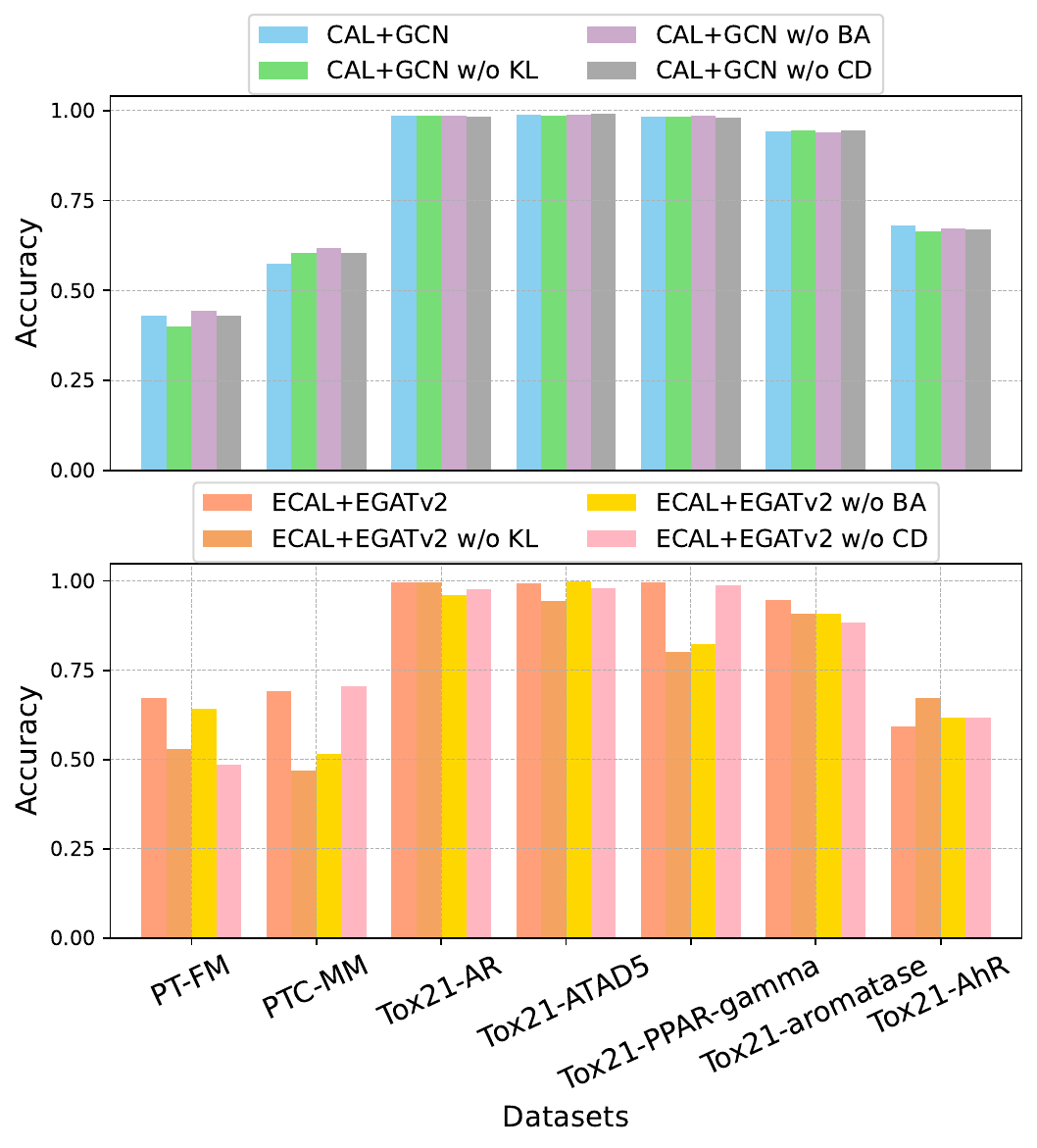}
    \vspace{-1em}
    \caption{The effect on OOD accuracy of removing causality detection loss terms, for both CAL-based and ECAL-based models.}
    \label{fig:ablation}
\end{figure}

To evaluate the impact of causal intervention techniques and highlight the importance of incorporating edge features in causal learning, we remove the causality detection loss terms in ECAL and CAL. Specifically, we compare removing the KL divergence term~\eqref{eqn:loss-kl} (KL), the backdoor adjustment term~\eqref{eqn:loss-ba} (BA), or both (CD). Results are shown in~\Cref{fig:ablation}. For CAL, which does not use edge features, removing these terms has little effect on accuracy under label imbalance. In contrast, ECAL with edge features improves test accuracy in such settings by leveraging causality detection loss terms.

\subsubsection{Adding noise to ablate edge features}
\begin{figure}[t!]
    \centering
    \includegraphics[width=0.95\linewidth]{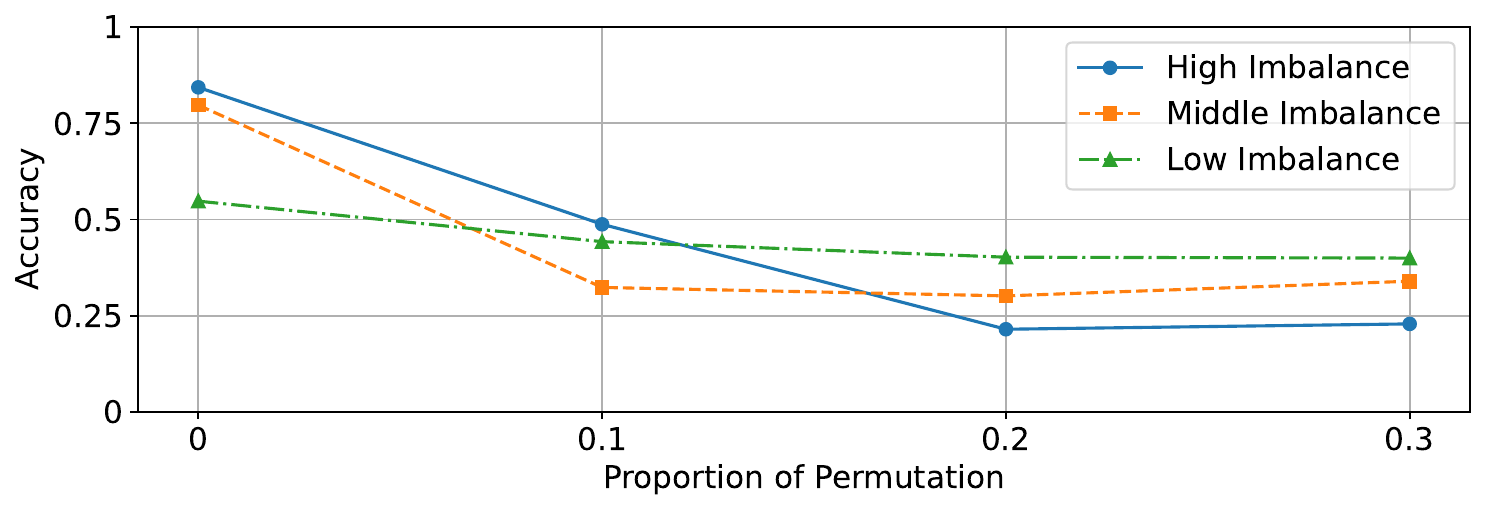}
    \vspace{-1em}
    \caption{OOD accuracy under different noise levels (proportion of permutation) and label imbalance levels on ogbg-molhiv.}
    \label{fig:noise}
    \vspace{-0.5em}
\end{figure}

In causal learning, features are considered non-causal if adding noise does not reduce OOD accuracy. 
To emphasize the causality of edge features, 
we randomly permute a proportion of edge feature vectors. 
As shown in \Cref{fig:noise}, adding noise consistently reduces OOD accuracy on ogbg-molhiv, confirming the causal significance of edge features.

\section{Conclusions}

In this paper, we proposed an edge-enhanced framework for causal attention mechanism, which applies edge-featured graph attention networks to better separate causal and non-causal subgraphs and extract the graph representation.
This new design extends the capabilities of causal attention mechanisms, offering a promising direction for tackling diverse graph-level tasks in real-world applications.
We validate the effectiveness of our proposed models on benchmark datasets (label imbalanced), including PTC, Tox21, and ogbg-molhiv; our proposed methods demonstrate superiority in label imbalance scenarios compared to existing methods and showcase the necessity of incorporating edge features.

\clearpage

\section*{Acknowledgments}
This work is supported by SF Tech. Yifan Chen is supported by the Research Grants Council (RGC) under grant \texttt{ECS-22303424}, and GuangDong Basic and Applied Basic Research Foundation through the general project grant.

\bibliographystyle{IEEEtran}
\bibliography{refs}

\end{document}